# Data generation using simulation technology to improve perception mechanism of autonomous vehicles

Minh Cao[1], Ramin Ramezani[2]

[1] Bioengineering Dept, University of California, Los Angeles, Los Angeles CA 90095, USA
[2] Computer Science Dept, University of California, Los Angeles, Los Angeles CA 90095, USA

**Abstract.** – Recent advancements in computer graphics technology allow more realistic rendering of car driving environments. They have enabled self-driving car simulators such as DeepGTA-V and CARLA (Car Learning to Act) to generate large amounts of synthetic data that can complement the existing real-world dataset in training autonomous car perception. Furthermore, since self-driving car simulators allow full control of the environments, they can generate dangerous driving scenarios that the real-world dataset lacks such as bad weather and accident scenarios. In this paper, we will demonstrate the effectiveness of combining data gathered from the real-world with data generated in simulated world to train perception system on object detection and localization task. We will also propose a multi-level deep learning perception framework that aims to emulate a human learning experience in which series of tasks from simple to more difficult ones are learnt in a certain domain. The autonomous car perceptron can learn from easy-to-drive scenarios to more challenging ones customized by simulation software.

## 1 Introduction

Autonomous vehicles require accurate and reliable perception systems to navigate through complex urban environments. Such systems normally rely on images from cameras coupled with 3D point cloud maps generated by lidar to gather information about texture, shape, and distance of objects in the surrounding environment. Interpreting raw data from these sensors and combining them to piece together a complete picture of the environment has been a subject of study for decades. With drastically increased computational power over the past few years and their accessibility, artificial neural networks, particularly building deep learning algorithms, have gained significant traction in a wide range of applications, including autonomous vehicles. Deep learning models are amongst the most successful solutions in building autonomous vehicle perception modules. However, training data driven techniques such as deep learning to perform perception, i.e., object recognition and localization, requires a large amount of annotated data. Traditionally, the training data is collected via a fleet of vehicles mounted with camera and lidar sensors driving around the location of interest. Such data needs to be semantically segmented, i.e., to generate the bounding box, pixel-level mask and 3D point cloud object level of every captured object within locations of interest. Objects' semantic segmentation is an exceedingly labor-intensive task in that it is primarily done manually by humans [1-2]. There are several studies [3-4] on machine assisted annotations but none are fully autonomous, and they all require humans' supervision to perform the final check on each individual sample data. Moreover, even if the annotation could be executed manually in a short period of time, severe shortage of training data covering dangerous driving scenarios would still pose a problem in developing fully autonomous algorithms; scenarios that may arise in driving in rough weather or terrain conditions such as rain, snow, rural areas during night or even accidents themselves. Datasets including dangerous scenarios are rare and difficult to collect which result in training algorithms on sets largely biased towards less risky scenarios. Therefore, balancing the training data by increasing high risk driving scenarios is of paramount importance in developing a robust autonomous vehicle perception system.

Due to recent advances in video game computer graphics, it is now possible to generate photo-realistic driving scenarios in a simulated virtual world. Additionally, since the game engines have complete knowledge of every object in the virtual world, they can be used to automate the semantic segmentation labeling process. Some studies have been leveraging this ability, especially from the Grand Theft Auto V (GTA-V) game, to generate synthetic training data to train autonomous vehicle perception systems. In [5], models trained on only video game driving images outperform models trained on images from the real-world Cityscapes dataset [6]. In [7] authors leverage the DeepGTAV engine [8] to simulate lidar sensors and to collect 3D point cloud data from the game world. The study indicates that complementing these data with real world KITTI dataset [9] improves their models' performance on segmentation tasks. The study demonstrates how to leverage the GTA-V's engine capability to generate camera images and lidar 3D point cloud data rapidly and automatically. The GTA-V game engine, however, lacks the ability of high degree customization, especially on weather and non-player object control, to be harnessed as a robust simulator for autonomous vehicles. High level of customization is necessary to simulate dangerous driving scenarios that are either not available or very rare in real-world dataset.

In this paper we aim to investigate leveraging an improvement on GTA-V engine - CARLA (Car Learning to Act) simulator [10]. CARLA is a simulation platform based on the open-source physics game, Unreal Engine 4 [11]. Being a platform specifically designed to aid autonomous driving development, CARLA not only provides full control on weather conditions and objects in the virtual world but also provides a simple Python API to collect data from many built-in sensors. The list of built-in sensors include camera, lidar, depth and most importantly semantic segmentation sensors that color-code every type of objects in the simulated world. With the ability to fully control the intricacies of the simulated world, we can simulate dangerous real-life driving scenarios such as snow, fog and heavy rain that are hardly presented in publicly available datasets. Moreover, we can recreate accident scenarios such as the 2018 Uber crash [12].

Armed with this new tool that enables us to design, generate and collect different driving scenarios in the form of training data, we propose a new framework for a multi-level perception system. This framework is inspired from human learning experience at school. In order to teach new concepts, the teacher may provide students with examples with increasing difficulties. This allows students to better grasp the core concepts from simple problems and combine them to solve more difficult problems. In the same way, we will generate training data with different levels of difficulty to train our perception system level by level. Each level will be associated with a unique classifier that will remain unchanged when moving to higher level. This way, the core concepts are captured at the lower-level classifiers, which allows higher level classifiers to focus on developing new techniques and to tackle more difficult scenarios. To the best of our knowledge, there has been no similar proposal of this type of framework on autonomous vehicle applications. This proposal is aimed to have the following **contributions**:

1. To demonstrate that supplemental synthetic data from the simulated world can improve performance of object recognition and localization compared to models that are solely built on real-word annotated training data.
2. To demonstrate the efficacy of simulating risky and dangerous driving conditions in improving the autonomous vehicle perception tasks which are primarily based on real world training data [17-19]. We aim to simulate rain, snow, darkness and car accidents and feed them to learning algorithms as training data.
3. To propose a framework for a multi-level learning model that emulates human learning experience at school by customizing the difficulty of synthetic data.

## 2    Methodology

This section describes our proposed approach to:

- generate synthetic training data with stereo camera images and 3D map lidar data points with computer-generated annotations,
- segment images of vehicles, pedestrians and cyclists, which are then used to improve object detection approaches from the work of [17-19] as well as to train our new proposed object detection framework.

### 2.1    Training data collection procedure

We leverage the virtual world from CARLA - an open-source simulator developed for urban driving scenarios - to recreate the driving environment encountered by autonomous vehicles. The environment details are captured through a set of sensor suites provided by the simulator: Lidar, RGB, depth and segmentation cameras. The sensors can be attached to vehicles at any location and facing direction. The simulator allows the gauging of different sensor configurations. Among CARLA's provided sensor suites, the segmentation camera stands out as the main advantage of the simulated training dataset compared to real world dataset. This sensor generates camera images with objects color-coded according to their type (car, pedestrian, traffic light, etc.). Essentially, this is a tool to guarantee "rapidly generating accurate ground truth in pixel level" that is required for many popular perception algorithms utilizing segmentation masks for training. Since the underlying game engine has perfect knowledge of all objects in the simulated world, it is possible to rapidly generate ground truth labels for other types of sensors as well (e.g., generating bounding boxes of all objects in the camera image as well as locations and heading directions of all objects hit by lidar rays).

Currently, Carla simulator does not provide tools to automatically generate localization information to label training data. Therefore, we built a preliminary tool to automatically generate object labels captured by the camera the same as the KITTI dataset format, which is the most commonly use dataset for perception models. We **propose** the following steps to generate labels:

1. Request information of nearby objects of interest. We recommend labeling cars and pedestrians in the radius of 120 meter. Bear in mind that although the lidar coverage is 200 meters, 120-meter coverage is far enough to capture comprehensive images and not too far to result in segmenting irrelevant ones.
2. Generate bounding box vertices for each object.
3. Project vertices of each object from world view to camera view and then to pixel view.
4. Determine if the pixel surrounding the projected pixel is occluded. This is done by comparing pixels' depth to corresponding depth from depth map image
5. If more than 2 vertices of the object are not occluded and the object pixel height is more than 25, the object label is generated according to KITTI format shown in Table 1.

**Table 1.** Label format of KITTI dataset [9]

| #Value | Name | Description |
|---|---|---|
| 1 | Type | Describes the type of object: 'Car', 'Van', 'Truck', 'Pedestrian', 'Person_sitting', 'Cyclist', 'Tram', 'Misc' or 'DontCare' |
| 1 | Truncated | Float from 0 (non-truncated) to 1 (truncated), where truncated refers to the object leaving image boundaries |
| 1 | Occluded | Integer (0,1,2,3) indicating occlusion state: 0 = fully visible, 1 = partly occluded, 2 = largely occluded, 3 = unknown |
| 1 | Alpha | Observation angle of object, ranging [-pi..pi] |
| 4 | Bbox | 2D bounding box of object in the image (0-based index): contains left, top, right, bottom pixel coordinates |
| 3 | Dimensions | 3D object dimensions: height, width, length (in meters) |
| 3 | Location | 3D object location x, y, z in camera coordinates (in meters) |
| 1 | Rotation_y | Rotation ry around Y-axis in camera coordinates [-pi..pi] |

## 2.2 Scenario generation

The main feature that sets CARLA simulator apart from GTA-V game engine used in [5][7] is the ability to gain control of all aspects and intricacies of the virtual world, most importantly objects and weather. One application of the weather control is the ability to improve autonomous vehicle perception by simulating roads under various weather conditions. Weather conditions have a dramatic effect towards both cameras and lidar sensors' perceptions. The most notable one is light source (angle of the sun and car headlights, etc.) and precipitation (rain and snow).

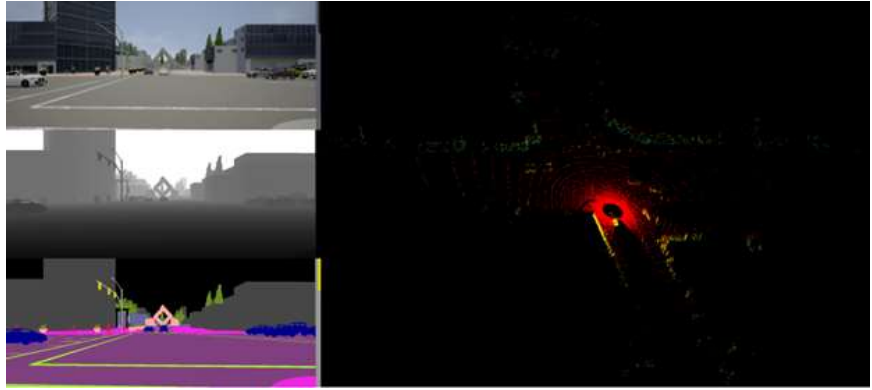

**Fig. 1.** Simulated data using CARLA. Image on the right is the lidar output (100k point sample), Top Left shows RGB Camera, Middle Left shows depth camera. Bottom Left is segmentation camera in which the presence of a car is indicated in blue, pedestrian in red, traffic light in yellow, road in purple and lane mark is indicated in green

The sun's angle will determine how shadow is cast in the environment and lens flare forms on camera image, both of which will hinder camera-based detection algorithm; the reduced contrast of objects falling in the shadow from their former state causes undesired image artefact. The effect of lens flare is even more dramatic at nighttime with artificial lights emitting from buildings, road reflectors and other car headlights.

While the light source only affects camera-based detection, precipitation affects both camera and lidar. Reflected light from rain puddles can confuse camera-based object recognition algorithms and may intensify lens flare effect. Raindrops and snowflakes, registered in lidar sensors, can reduce the resolution of real objects in 3D cloud point map and can also generate more noisy objects. Available annotated training datasets on such weather conditions are few despite being highly prone to include crash incidents. Weather conditions such as blizzards or stormy rain are nonexistent due to their seldom occurrence in real life, not to mention collecting those conditions puts test cars in danger. It is therefore ideal to supplement traditionally collected datasets with synthetically generated ones.

CARLA simulator is currently capable of changing the sun's position, cloud cover, precipitation and the way rain puddles on streets. Artefacts caused by these conditions such as reflection, shadow and lens flare, can be simulated on camera sensors to a large extent. Figure 2 depicts various scenarios we generated using CARLA with various weather conditions.

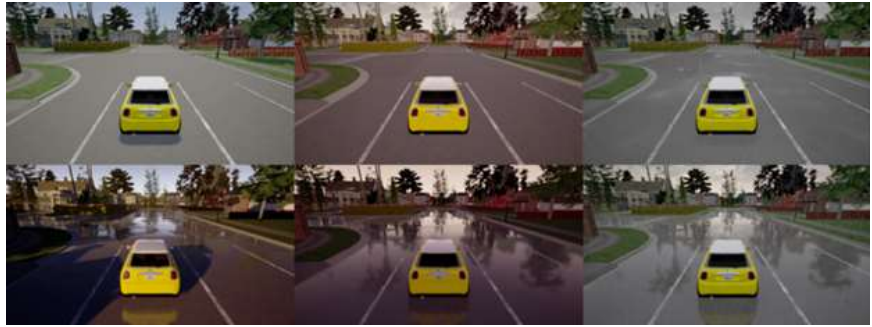

**Fig. 2.** Simulated Scenarios of different weather condition: clear at noon (top left), cloudy sunset (top middle), light rain (top right), hard rain (bottom right), after rain cloudy sunset (bottom middle), after rain clear noon (bottom right)

### 2.3 Generating Synthetic Data: Improving the performance of perception system using supplemental synthetic training data generated by simulator and enhanced by Style Transfer Conditional Generative Adversarial Network

In order to show that training data generated from the simulation is comparable to traditional datasets, we feed our training data to pretrained object localization algorithms (Yolo [20] and MobileNetV2 [21]) from other studies and evaluate the performance. The quality of the images generated from the CARLA simulator can be further enhanced to be more realistic using the style transfer conditional generative adversarial network (cGAN). In a normal generative adversarial network (GAN), the model is provided with samples of the wanted product, e.g., images from the car's front-camera. From those samples, the trained model will learn how to generate similar results given the random input. This alone is not helpful for our purpose since we want the generated images to not only look realistic but also make sense in the physical world, e.g., car on street lane, not on sidewalk. To improve this, cGAN takes in extra conditions of the desired output on top of the random input in order to generate a more directed result [22]. The conditions here are usually the wanted class label. In our use case, we will feed the model with the semantic labeled images that are generated alongside with the synthetic image in order to construct the realistic version of that image. [23, 24] have shown the feasibility of this technique on generating photorealistic synthetic car dash camera footage. In our proposal, we utilize this technique to further enhance our object detection algorithm. We will train our cGAN model on the Cityscape dataset [6] with the camera images as ground truth and the semantic segmentation images as the condition. Afterward, we will use the trained model to generate the realistic images from the semantic segmentation images generated by the CARLA simulator. These enhanced images will be used to supplement our object detection model.

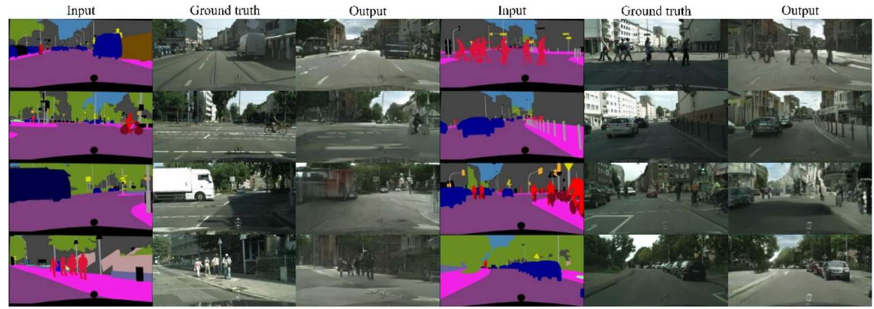

**Fig. 3.** Example of translating semantic image to photo realistic image using cGAN. Input is the semantic segmentation image. Ground truth is the actual camera image. Output is the result of our implementation of [23] on the Cityscape dataset [6].

## 3      Object Recognition Experimentations and Result

### 3.1    Dangerous and rough driving weather and terrain conditions

The goal is to show the simulated training dataset with different weather conditions can help improve the object detection algorithm even though such conditions are not presented in the original real-world dataset. The set up for this dataset is identical to the one discussed above with just a minor difference: Since the KITTI real world dataset does not contain the validation set with different weather conditions, the Waymo open dataset will be used as the "Test dataset". Waymo open dataset contains fully annotated camera and lidar training samples under different driving scenarios including raining, cloudy and sunset captured in real life. In summary, we use simulated dataset along with KITTI for training purposes and Waymo as the test set. By evaluating the performance of selected object detection algorithms with the additional synthetic training data consisting of varying weather conditions, we will investigate if the extra training data can contribute to the improvement of the learning algorithm. Figure 4 shows the result of using Yolo and MobileNetV2, trained on real-world dataset as well combined with our simulated data generated using CARLA. The test bench is Waymo dataset. As shown in the figure, the confidence intervals increase in both combined scenarios and at night condition, MobileNetV2 model performs better in detecting both cars when trained on real and simulated dataset combined.

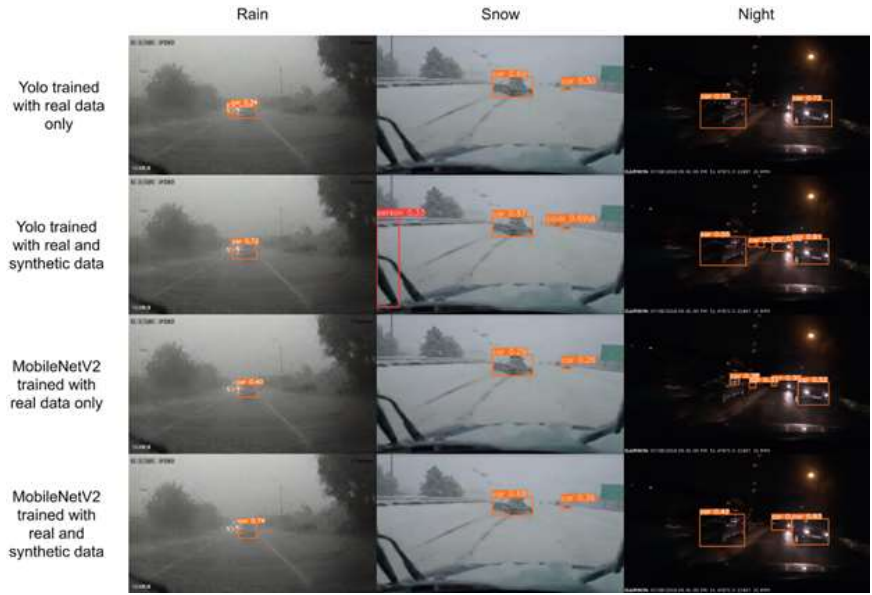

**Fig. 4.** Example of objection detection ROI result using Yolo and MobileNetV2 trained on KITTI dataset under different weather conditions.

### 3.2 Accident scenarios

Training data on car accident scenarios is of paramount importance for autonomous vehicles development. Nonetheless, no publicly available dataset (as far as authors knowledge) exists that contains data captured from accidents. There could be 3 reasons for this: a) car accident are rare compared to normal driving scenarios, b) vehicles equipped with sensors to collect training data have normally extra cautious drivers (in fact there is only one instance of fatal crash for car equipped with lidar: 2018 Uber Case) and c) Even when the vehicles capture an accident scenario (e.g. tesla and uber crash), the company behind them is reluctant to release the complete data to public. In every Tesla's crash, the company has only provided its own internal analysis to investigators without disclosing any camera footage of the crash to the public. The only crash footage video released to the public is from the 2018 Uber crash in Arizona.

Fortunately, even though there is no data on crash scenarios, descriptions of how the accidents occur are available publicly on police investigation documents from the National Transportation Safety Board. From these descriptions, similar accident scenarios can be generated from the CARLA simulator to collect training data. Since these scenarios can be generated in large scale, they can contribute in solving the imbalance dataset problem that originated from the essential bias towards normal driving scenarios. The generated dataset from accident scenarios may improve object localization and can ultimately result in early prediction of possible incidents.

There is one issue with this approach worth highlighting. The performance of the new model may not be well gauged due to the lack of publicly available datasets on accidents. Using dash camera footage of personal cars however may mitigate the problem. Those footages normally cover moments leading to accidents and are widely available on Youtube. The footages can provide a source of accident data for model testing purpose. Although, using dash-cams are not also without a challenge. Dash-cam videos are 2-D and can be used to evaluate the object recognition performance of the proposed model. We investigated dash cam videos covering the following accident scenarios:

1. car from the side suddenly swirl to the front [13-14]
2. pedestrian or car at night occluded in the shadow shown in [15] at timestamps 0:34 and 2:17 and [16] at 0:28 and 6:52.

These two types of accident scenarios share common characteristics: a) there is a sudden increase in the number of visible pixels and b) early and consistent detection between frames is crucial to avoid accidents. The setup to evaluate the improvement of extra dataset is as follows:

1. Generating two types of accident scenarios in CARLA with characteristic as described above and collect camera data with bounding box label on it
2. Training selected camera-based object localization from other works at [17-19] with varying amount of synthetic data: normal driving only/normal driving + accident scenario
3. Generating evaluation dataset:
   a. extracting accidents video footages captured by dash cameras available on Youtube into 150 image frames, cutting each to 10 seconds
   b. manual annotation with bounding boxes of cars and pedestrians in each frame.
4. Evaluating the trained model from step 2 on the dataset from step 3. The performance metrics we aim to consider are a) the frame from which the first detection of the car/pedestrian happens, b) the average percentage of total frame from which the car/pedestrian is detected and c) the average consecutive detections of car/pedestrians.

Figure 5 shows our implementation of Yolo and MobileNet models trained on simulated data combined with real data, tested on Uber crash scenario [12]. It is clear that both the Yolo and MobileNetV2 algorithms trained on the combined dataset detected pedestrians in earlier frames compared to their counterparts trained only on real data.

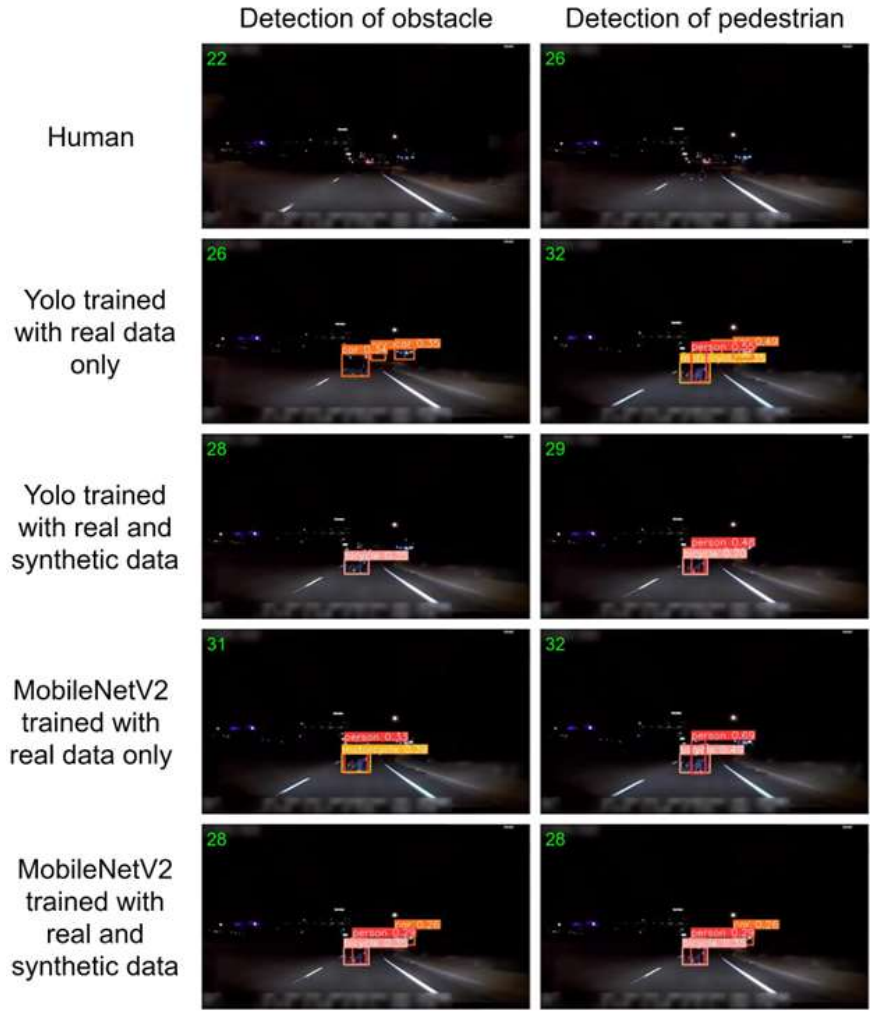

**Fig. 5.** Frame by frame objection detection ROI using Yolo and Mobile Net model on Uber car crash video. Numbers on each image show the frame number in the video.

## 4 Multi-stage deep learning perception framework

To further improve our existing model, we propose a multi-level deep learning framework for autonomous vehicle object recognition and localization. This model will leverage the combination of the simulated and real-world data. There are two main goals we want to achieve with this framework: 1) to learn the common objects' features between the simulated world and real-world 2) to emulate human learning processes in which humans learn series of tasks from easy-to-solve to more challenging ones in a particular domain.

### 4.1 Learning common features between simulated world and real world

There are two main characteristics that make simulated and real-world datasets different:

1. Objects generated by simulator programs will lack detailed texture compared to real world objects. This will make features extracted from the simulated world to have different characteristics from the real-world when using identical feature extracting networks. In other words, due to variations in texture, albeit insignificant, feature extracting networks may form different perceptions of the same objects. To rectify this problem, we aim to learn the commonalities (common features) between identical objects generated by simulators versus the ones perceived in the real-world.
2. Training dataset from simulated world can be generated much more rapidly than real-world dataset. For instance, we were able to generate the equivalent of KITTI camera and lidar dataset in 12 hours using a computer with 8 core 3.6 GHz CPU with RTX 2080 GPU. In other words, if we train the perception model on the training dataset that is dominant with simulated data, the model may likely be biased towards the features that are unique to the simulated world. We could limit the amount of simulated data in the ultimate dataset to avoid this situation.

In order to compensate for the different characteristics of synthetic and real-world datasets, we arrange our datasets as follows: Simulated dataset will form the training dataset while the real-world dataset will be considered as the validation set. Using this arrangement, the parameters of the training model will be fitted by the simulated data while its hyperparameter is fitted by the real-world data. This arrangement allows us to evaluate the effectiveness of a model in translating knowledge gained from simulated world toward the real-world. The rationality behind this is that even though the texture of the object in the simulated world is not identical to the one in the real world, humans can still identify objects from the simulated world correctly because there is a shared general characteristic of the object from which the simulated world was produced from (e.g., general shape, size). This arrangement will allow us to focus our model on those general features as well as leverage the large dataset generated from the simulated world. From now on, we refer to this arrangement as Simulated Training Real Validation (STRV).

### 4.2 Multi-stage deep learning perception framework

We will use the dataset arrangement discussed above for our framework. The general architecture of the framework is illustrated in Figure 6.

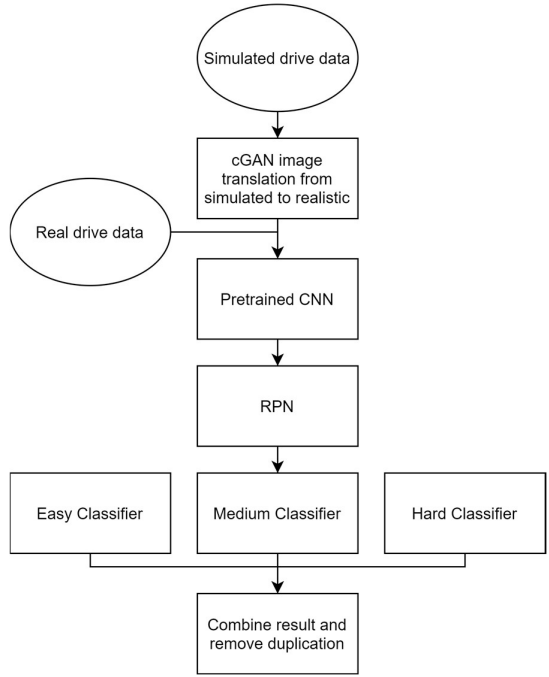

**Fig. 6.** Multi-Stage Deep Learning framework

1. Using the pretrained Convolutional Neural Network (CNN) on commonly used datasets, ImageNet and Coco, to generate features from input images. The generated features will be fed to the Region Proposal Network (RPN), which will subsequently generate candidate bounding boxes that contain objects of interest.
2. Training the RPN using the STRV arrangement. This makes the RPN focus on extracting features that are shared between real world and simulated data.
3. The main contribution of this section is the classifier layer. We want our three different classifiers to represent 3 different stages of learning: easy, medium and hard. Each classifier is trained with a dataset corresponding to its difficulty stage. For example, the easy classifier will be trained on a dataset with a small number of objects that are all fully visible while the hard classifier will be trained on a dataset composed of large number of objects that are being occluded.

To enable this framework, we need to consider two steps: a training dataset with defined difficulty and a process to enforce additive learning.

### 4.3   Training dataset with defined difficulty

We aim to concentrate on the following characteristics of an image within which objects are difficult to be correctly classified: occlusion, number of objects, bounding box pixel area. The simulated and real-world data will be categorized based on these characteristics to make datasets with different levels of difficulty.

**Table 2.** Criteria to categorize training data into different levels of difficulty

| Difficulty | Occlusion level | Number of objects | Bounding box pixel area |
|---|---|---|---|
| Easy | <20 % | <3 | >400 |
| Medium | 20 - 50 % | 3 - 6 | 100 - 400 |
| Hard | >50% | >6 | 25 - 100 |

This categorization process can be done rapidly in the simulated portion since we have access to the image rendering engine as well as the state of all objects in the image to infer all the necessary characteristics of the sample. For the real world portion, although the number of objects and bounding box area is available in the label of each dataset, the label for the level of occlusion will require manual work. We will do experiments to determine the optimal proportion of simulated real-world data under STVR dataset arrangement. The higher this proportion is, the less manual work to label real world data will be required.

### 4.4  Process to enforce additive learning.

The process of training with additive learning is as followed:

1. The general architecture will be trained with the easy dataset as described above. There is only one classifier (easy) in this step. The weight of the CNN and RPN layer will be frozen. Only the weight of the classifier will be modified since we are only interested in training the classifier to recognize objects included in the proposed bounding box from the RPN layer. The classifier will come from the chosen sample of other camera-based object detection work. The goal of this step is to allow the classifier to learn the core basic concepts of object classification without the effect of occlusion, small size and crowdedness.
2. Repeat the same process as step one on the medium dataset. This time, an additional classifier (medium) will be added. This additional classifier will be the only one with weight being modified. The detection result of the two classifiers are combined. Detection results of the medium classifier with the same class and in close proximity to the results from the easy classifier are discarded. Because of this arrangement, the basic concept to classify objects learned from the easy classifier persist and take higher priority. This will force the medium classifier to come up with new techniques to detect objects that are missed by the easy classifiers due to moderate effect of occlusion, small size and crowdedness.
3. Repeat the same process as step two on the hard dataset with another additional classifier (hard). The training logic is the same as above. This time, the hard classifier is forced to come up with new techniques to detect objects that are missed by both the easy and medium classifiers due to severe effects of occlusion, small size and crowdedness.

Our proposed framework is not limited to 3 classifiers only. It can be generalized to have any number of classifiers corresponding to levels of difficulty. The flexibility in the framework provides a platform that enables users to choose a wide range of features; features that ultimately define the difficulty of objection detection.

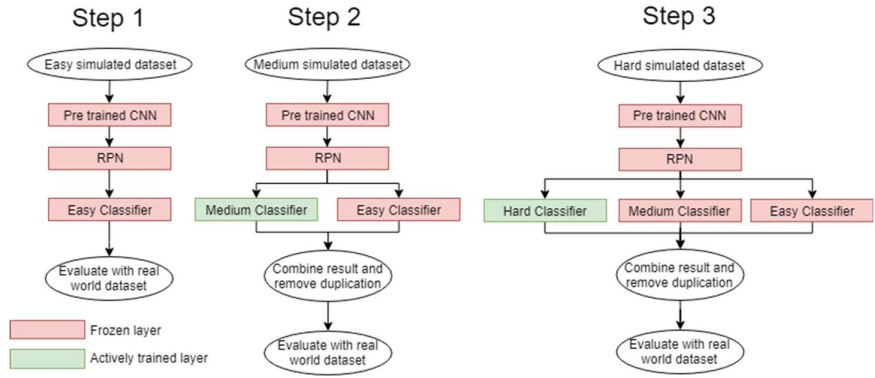

**Fig. 7.** Training process of the classifier layer of the proposed architecture. Only green box modules' weights are modified during training process. The red box modules' weights are unchanged